\pdfoutput=1

\documentclass[11pt]{article}

\usepackage{ACL2023}

\usepackage{times}
\usepackage{latexsym}

\usepackage[T1]{fontenc}

\usepackage[utf8]{inputenc}

\usepackage{microtype}

\usepackage{inconsolata}

\usepackage{comment}
\usepackage{graphicx,multirow}
\usepackage{xcolor}
\definecolor{myblue}{RGB}{79, 134, 236}
\definecolor{myred}{RGB}{235, 50, 35}

\newcommand{\shortpar}[1]{\noindent \textbf{#1}}

\newcommand{\ignore}[1]{}

\title{Grounding in Open-domain Conversational Question Answering}
\title{Trust me bro: Faithfulness and Trust in Open-domain Conversational  Question Answering}
\title{People trust Stochastic Parrots -- even when they hallucinate}
\title{People trust Stochastic Parrots. Faithfulness and Trust in Open-domain Conversational  Question Answering}

\title{The Dangers of trusting Stochastic Parrots: Faithfulness and Trust in Open-domain Conversational  Question Answering}

\author{Sabrina Chiesurin*~~Dimitris 
Dimakopoulos*~~Marco Antonio Sobrevilla Cabezudo \\ {\bf Arash Eshghi~~Ioannis Papaioannou~~Verena Rieser$\dagger$~~Ioannis Konstas}  \\
  Alana AI\\  
\texttt{hello@alanaai.com} \\
  }

\begin{document}
\maketitle
\def\thefootnote{*}\footnotetext{Equal Contribution.}
\def\thefootnote{$\dagger$}\footnotetext{Now at Google DeepMind.}
\def\thefootnote{\arabic{footnote}}
\begin{abstract}
Large language models are known to produce output which  sounds fluent and convincing, but is also often wrong, e.g.\ ``unfaithful" with respect to a rationale as retrieved from a knowledge base. In this paper, we 
show that task-based systems which exhibit certain advanced linguistic dialog behaviors, such as lexical alignment (repeating what the user said), are in fact preferred and trusted more, whereas other phenomena, such as pronouns and ellipsis are dis-preferred. 
We use open-domain question answering systems as our test-bed for task based dialog generation and compare several open- and closed-book models. Our results highlight the danger of systems that 
appear to be trustworthy by parroting user input while providing an unfaithful response. 
\end{abstract}

\section{Introduction}

With the advent of large language models (LLM), Question Answering Systems have become open-domain and conversational, meaning that they are  able to generate fluent and informative responses to questions about nearly any topic and over several turns \cite{adlakha-etal-2022-topiocqa}. However, these systems are also known to produce factually incorrect statements, commonly referred to as \textit{hallucinations} \cite{rashkin-etal-2021-increasing, dziri-etal-2022-origin}.
These two properties taken together require the system as well as the user to ensure that they mutually understand each other -- a process also known as {\em conversational grounding} \cite{clark1991grounding}.

\begin{figure}[t]
    \centering
    \includegraphics[width=1.0\columnwidth]{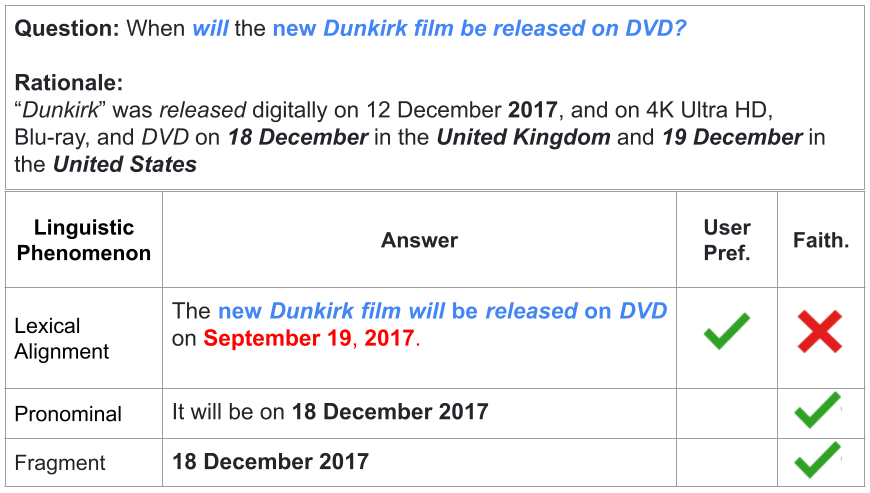}
    \caption{Responses with different forms of conversational linguistic phenomena and token grounding: \textbf{\textcolor{myblue}{Blue}} indicates tokens from the question are repeated in the response ({\em lexically aligned}). \textbf{Bold} corresponds to content tokens in the response \textit{grounded} in the knowledge source; 
    \textbf{\textcolor{myred}{red}} tokens are hallucinations, i.e., not \textit{faithful} to the dialogue and rationale. The last two columns indicate user preference and faithfulness, respectively.} 
    \label{fig::faithfullness-grounding}
\end{figure}

Empirical studies of dialogue have shown that people use different kinds of context-dependent linguistic behavior to indicate grounding, including use of fragments, ellipsis and pronominal reference \cite{fernandez-ginzburg-2002-non-sentential, https://doi.org/10.1111/cogs.12225}. Other studies show that lexical alignment in a response, i.e.\  repeating and adopting the interlocutor's lexical items \cite{pickering2004toward,branigan2010linguistic}, can play a similar role, see examples in Figure~\ref{fig::faithfullness-grounding}.

\begin{table*}[]
    \small
    \centering
    \begin{tabular}{r|cccccccccc}\hline
        \textbf{Models} & \textbf{Length} ($\mu$) & \multicolumn{3}{c}{\textbf{Structure} (\%)} & \multicolumn{3}{c}{\textbf{Align}} & \textbf{Pron} (\%) \\\hline 
        && \textbf{Frag} & \textbf{Short} & \textbf{Long} & \textbf{P} & \textbf{R} & \textbf{F1} & \\\hline
        DPR+FiD & 9.1 & 64.6 & 33.4 & 2.0 & 6.1 & 8.7 & 6.2 & 23.1 \\
        DPR+GPT-3 & 24.4 & 13.9 & 56.7 & 29.4 & 14.8 & 37.0 & 19.5 & 10.4 \\
        GPT-3 & 20.1 & 12.5 & 55.8 & 31.7 & 18.3 & 38.0 & 22.9 & 12.1 \\\hline
        Human  & 11.2 & 57.0 & 36.2 & 6.8 & 6.6 & 10.8 & 7.2 & 18.7 \\\hline
    \end{tabular}
    \caption{Linguistic phenomena of responses for different models on the development set of TopiOCQA.}
    \label{tab:dialogue-phenomena}
\end{table*}
%
There is initial evidence in related fields that generating grounding phenomena will lead the user to trust the system more, such as conversational assistants for educational \cite{Linnemann2018:lexicalAlignmentTrust} and medical applications 
\cite{Bickmore2021:medicalTrust} as well as in the field of
HRI \cite{Bossens2022:HRIgrounding}. At the same time, we argue that systems that exhibit more grounding behavior are not necessarily more faithful to the dialogue and input rationale, which can lead to unjustified trust. 

In order to explore these hypotheses,  we first analyze conversational grounding phenomena via automatic annotation of linguistic properties for open-domain QA. We consider responses generated by different  GPT-3 variants \cite{NEURIPS2020_1457c0d6}, and state-of-the-art Retrieve-and-Generate models on the TopiOCQA development set \cite{adlakha-etal-2022-topiocqa}. 
We evaluate the performance of models via several automatic surface-level, and semantic-based metrics against multiple references and a chosen rationale from a gold Wikipedia passage. 
Given current limitations of automatic metrics, we  annotate a subset of responses according to their plausibility, groundedness to the input source and faithfulness to the dialogue and input source \textit{at the same time}. We also elicited a human preference task among the responses of each model.
Finally, we conduct a series of human evaluation experiments where we provide responses to questions controlling for each of the linguistic phenomena under examination, and ask users to choose the one they perceive as more trustworthy.
Our findings are summarised as follows:
\begin{itemize}
    \item GPT-3 variants are generally more verbose and more lexically aligned to the question. In contrast, the human-authored responses in TopiOCQA are more elliptical and contain more pronominals. Unsurprisingly, the fine-tuned model emulates this behavior.
    \item GPT-3 variants are 
    less faithful according to expert human annotations and the majority of automatic metrics. 
    \item Surprisingly, users prefer open-book \mbox{GPT-3} over the fine-tuned model although half of the time the preferred responses were unfaithful.
    \item Users trusted responses with high lexical alignment significantly more, whereas the effect was the opposite for elliptical responses, and answers containing pronominals.
\end{itemize}

\section{Conversational Grounding Analysis}
\subsection{Dataset and Models}
\shortpar{Dataset} We use the development set of TopiOCQA comprising 205 information-seeking dialogues (2514 turns)\footnote{A manual analysis of the dataset revealed that the linguistic phenomena under scrutiny are almost exclusively present.}. 

\shortpar{Models}
We test a variety of models under two different settings. In the {\em closed-book} setting models have no access to domain-specific information other than what is stored in their own parameters; in the {\em open-book} setting models can leverage a set of relevant documents provided by the retriever. 

For the open-book setting we used a fine-tuned Dense Passage Retriever (DPR; \citealp{karpukhin-etal-2020-dense}) as the retriever and experimented with two different readers: Fusion in Decoder (FiD; \citealp{izacard-grave-2021-leveraging}) fine-tuned  on TopiOCQA,
and GPT-3 \cite{NEURIPS2020_1457c0d6}\footnote{We used \texttt{davinci-003} in all our experiments.}, where we concatenate passages returned from DPR with the dialogue context and use them as conversational prompt.
For closed-book similar to \citet{adlakha-etal-2022-topiocqa} we also use GPT-3, where the dialogue context is concatenated into a conversational prompt.

Notably, we could have also tuned GPT-3 either via prompt engineering or fine-tuning\footnote{Fine-tuning GPT-3 would entail several rounds of hyperparameter tuning increasing the cost of the experiments.} so that it resembles the distribution of the target dataset. We decided against this for two reasons: firstly, the amount of engineering required would go beyond the focused scope of this work; second using vanilla GPT-3 variants is as close as possible to an ecologically valid scenario. For example, it is similar to how an end-user would be exposed to an LLM via a search engine, or a chat interface without any direct control of its prompt. 

\begin{table*}[]
    \small
    \centering
    \scalebox{0.9}{
    \begin{tabular}{r|cccc|cc@{\,}c|c@{\,}cc}\hline
        & \textbf{F1} $\uparrow$ & \textbf{EM}  $\uparrow$ & \textbf{BLEU}  $\uparrow$  & \textbf{ROUGE}  $\uparrow$ & \textbf{BERT}  $\uparrow$ & \textbf{K-F1}  $\uparrow$ & \textbf{K-F1++}  $\uparrow$ & \textbf{Critic} $\downarrow$ & \multicolumn{2}{c}{\textbf{Q$^2$}}\\
        \textbf{Models} & & & & & & & & &  \textbf{F1} $\uparrow$ & \textbf{NLI}  $\uparrow$ \\\hline
         DPR+FiD & \textbf{55.3} & \textbf{33.0} & \textbf{44.74} & \textbf{56.3} & 0.79 & 21.3 & 19.0 & \textbf{55.9} & \textbf{32.8} & \textbf{35.9}\\
         DPR+GPT-3 & 37.4 & 5.9 & 20.02 & 39.0 & \textbf{0.81} & \textbf{28.4} & \textbf{22.6} & 63.2 & 26.5 & 29.8 \\
        GPT-3 & 33.9 & 6.8 & 12.71 & 36.4 & 0.80 & 20.2 & 15.7 & 59.2 & 19.9 & 24.3 \\\hline
        Human  & 70.1 & 40.2 & 58.63 & 70.8 & 0.83 & 33.0 & 29.3 & 20.7 & 59.9 & 63.6\\\hline
    \end{tabular}}
    \caption{Model performance using automatic metrics on the development set of TopiOCQA.}
    \label{tab:faithfulness-automatic}
\end{table*}
\subsection{Dialogue Phenomena}\label{sec:dialogue-phenomena}
We automatically annotate the following linguistic properties of responses:

\shortpar{Lexical Alignment} is approximated based on  unigram overlap between the response and corresponding question, i.e.\ the system repeating the same words as the user. This typically serves the purpose of implicitly confirming what was understood in task-based dialog. 
We compute the precision (P), recall (R) and F1. Figure~\ref{fig::faithfullness-grounding} shows a response that lexically aligns to the question.

\shortpar{Syntactic Form} We define three categories according to the syntactic structure, based on  the constituency tree\footnote{We used Stanza \cite{QiEtAl2020}.}:

\begin{itemize}
    \item \textit{short responses} comprise a single sentence with the tree's root being either a simple declarative clause (S), or a declarative sentence with subject-aux inversion (SINV); 
    see the first two responses in Figure~\ref{fig::faithfullness-grounding}.
    \item \textit{fragments} comprise an elliptic sentence, with its syntactic root not identified as either S or SINV; see last response in Figure~\ref{fig::faithfullness-grounding}.
    \item \textit{long-form responses} are multi-sentence answers, which are rarely occurring. This is probably due to the conversational nature of TopiOCQA where complex questions are broken down into simpler ones across a dialogue.
\end{itemize}


\shortpar{Pronominals} We identify the existence (or not) of a pronoun in a sentence in subject, or direct object position according to its dependency tree, 
e.g., ``\textit{It}" in the second response of Figure~\ref{fig::faithfullness-grounding}. 

Table~\ref{tab:dialogue-phenomena} summarizes the statistics of linguistic phenomena 
found in models and human responses. Note that GPT-3 variants produce more verbose, sentential and lexically aligned responses with the questions (see Recall column). In contrast, the fine-tuned model (DPR+FiD) generates shorter fragmented responses with more pronominals. This is expected as it follows the distribution of human responses, unlike the GPT-3 variants that 
have a very limited conditioning on the target distribution via the dialogue context getting encoded in the prompt.  

\section{Study of Faithfulness}


\shortpar{Faithfulness Definition} We extend the definition by \citet{adlakha-etal-2022-topiocqa} to consider faithfulness both wrt the \textit{dialogue} and rationale: 

\vspace{2ex}
\noindent \textit{Given a dialogue history} $\mathcal{H}=(u_1,...,u_{n-1})$ \textit{and knowledge} $\mathcal{K}=(k_1,...,k_j)$ \textit{at turn} $n$\textit{, we say that} utterance $u_n$ \textit{is faithful with respect to} $\mathcal{K}$ \textit{and} $\mathcal{H}$ \textit{iff} $\exists {\Gamma}_n$ \textit{such that} ${\Gamma}_n \models u_n \land E(\mathcal{H}, u_n)\neq \emptyset$, \textit{where} $\models$ \textit{denotes semantic consequence,} ${\Gamma}_n$ \textit{is a non-empty subset of }$\mathcal{K}$ \textit{and E is the explicature of} $u_n$ \textit{in context }$\mathcal{H}$ as defined in ~\cite{rashkin2021measuring}.


\subsection{Automatic Evaluation}
We first employ a wide range of automatic metrics to assess model performance grouped according to their similarity to a gold (human) reference (\textit{reference-based}), or their faithfulness to the provided knowledge $\mathcal{K}$ (\textit{reference-less}). 



\shortpar{Reference-based metrics} Following  \citet{adlakha-etal-2022-topiocqa} and \citet{https://doi.org/10.48550/arxiv.2204.10757}, we report F1 score, Exact Match (EM), BLEU \cite{PapineniEtAl2002} and ROUGE \cite{Lin2004}. These measure the overlap-based similarity between the generated response and the gold answer\footnote{Note that results for Human don’t go up to 100\% as each output is compared with 3 additional human annotations.}.

\shortpar{Reference-less token-level metrics} Similar to \citet{https://doi.org/10.48550/arxiv.2204.10757} and  \citet{shuster2021retrieval}, we report BERTScore (BERT) \cite{zhang2019bertscore}, and \mbox{Knowledge-F1} (K-F1). Notably, the latter calculates the unigram overlap between the response and a knowledge snippet $\mathcal{K}$, providing a verbatim measure of grounding to the input source.

We propose K-F1++, a variant of K-F1, that captures only the novel information in the generated response and discounts any lexical alignment to the question:  it calculates the unigram overlap between the response and $\mathcal{K}$, after \textit{subtracting} any tokens appearing in the question from the response.




 
 \shortpar{Reference-less entailment metrics} We report \textit{Critic} \cite{https://doi.org/10.48550/arxiv.2204.10757}, a dialogue-trained classifier determining if a response follows from a given snippet $\mathcal{K}$, and $Q^2$ \cite{honovich-etal-2021-q2}, which measures faithfulness via question answering.
 



\subsection{Human evaluation studies}
Similar to \citet{glaese2022improving}, \citet{bai2022training} and \citet{thoppilan2022lamda}, we conducted a human evaluation to assess the faithfulness of given responses, followed by a human evaluation study to collect human preferences when presented with two possible responses to an existing conversation.

\shortpar{Faithfulness Judgment task} Annotators are required to judge the plausibility of a response given the dialogue, the relevance of the gold passage to answer the question, and the faithfulness of the response given the dialogue and the gold passage. 
In more detail, we consider the response to be grounded when it (or a paraphrase of it) is found in the document. We consider a response to be faithful if, in addition to being grounded, it answers the question and follows from the dialogue. For example, given i) a conversation about European countries, ii) a document about European capitals, iii) a query ``\textit{What is the capital of Spain?}'', and iv) the response ``\textit{Castellano}'', if ``\textit{Castellano}'' is in the document, the response is grounded. However, it is not faithful with respect to the dialogue as it does not correctly answer the question.
Two annotators\footnote{The annotators comprise a hired annotator and one of the co-authors. Quality was ensured via multiple rounds of pilot annotations, until all disagreements were 
resolved.} completed the annotation for each model on 500 instances from TopiOCQA. 
    
\shortpar{Preference task} Annotators are provided with a question, the previous dialogue and the gold passage that contains the answer, and are required to select their preferred response given two options. These are between a baseline model (DPR+FiD) and a model variant; they can also select both or none. We take a sample of 250 faithful and unfaithful instances from the previous task.


\subsection{Results}
Table~\ref{tab:faithfulness-automatic} summarizes the automatic metrics. Baseline DPR+FiD outperforms the GPT-3 variants in all \textit{reference-based} metrics. This is somewhat expected since the former is fine-tuned on the \mbox{TopiOCQA} dataset, whereas GPT-3 --despite being a much larger model-- is evaluated in a zero-shot fashion. Surprisingly, DPR+GPT-3 outperforms the baseline in most \textit{reference-less} metrics. 

Interestingly, the absolute difference between K-F1 and K-F1++ with respect to the baseline (2.3\%) is significantly smaller than that of the GPT-3 variants (5.8\%, and 4.5\%, respectively). This is probably due to the latter being more lexically aligned to the user question than the baseline (see Table~\ref{tab:dialogue-phenomena}), hence there are more overlapping tokens removed when computing K-F1++. Nevertheless, the GPT-3 variants maintain superior knowledge-grounding scores even based on the stricter K-F1++.

Table~\ref{tab:faithfulness-human} paints a different story to the reference-less metrics: although all responses are regarded mostly plausible continuations to the dialogue, the GPT-3 variants (with the closed-book scoring worst) produce outputs that are less grounded and more unfaithful compared to DPR+FiD. We observed often the inclusion of extra information that could \textit{potentially} be true but still not faithful to the input source. We leave fact checking of such extrinsic hallucinations to future work. 

The most striking result according to the Preference task (Table~\ref{tab:preferences}) is that annotators preferred unfaithful responses over faithful ones, or rejected both options, even though they had access to the gold passage. 
DPR+GPT-3 overall was preferred 70\% of times, with almost half preferences being towards unfaithful responses (48\%). Similarly, \mbox{GPT-3} was preferred 45\% of the time with 66\% of preferences being unfaithful. Again this supports our hypothesis that high lexical alignment has a great influence on users' choices, often bypassing the need to judge the accuracy of the response. 
%
Appendix~\ref{app:additional_human_results} contains additional results on computing majority agreement per item among the 5 annotators for the Preference Task and a qualitative analysis of provided feedback. 

\begin{table}[]
    \small
    \centering
    \begin{tabular}{r|ccc}\hline
        \textbf{Models} & \textbf{Plaus.} & \textbf{Ground.} & \textbf{Faith.}\\\hline         
        DPR+FiD & 97.2 & \textbf{62.4} & \textbf{57.8} \\
        DPR+GPT-3 & \textbf{100.0} & 46.2 & 39.6 \\
        GPT-3 & 91.6 & 22.6 & 22.0 \\\hline
        Human  & 99.8 & 98.6 & 93.0 \\\hline
    \end{tabular}
    \caption{Faithfulness Judgement Task carried out by 2 expert annotators on a sample of 500 instances.}
    \label{tab:faithfulness-human}
\end{table}

\begin{table}[]
    \small
    \centering
    \begin{tabular}{@{}r|ccc}\hline
        \textbf{Model} &  \multicolumn{3}{c}{\textbf{Preferences}} \\ 
          & \textbf{All (\#)}& \textbf{Faith. (\#)}& \textbf{Unfaith. (\#)}\\\hline
        DPR+FiD & 33\% (417) & 85\% (354) & 15\% (63) \\
        None & 12\% (153) & - & - \\
        DPR+GPT-3 & \textbf{70\% (883)$\dagger$} & 52\% (459) & 48\% (424) \\\hline\hline
        DPR+FiD  & 43\% (539) & 84\% (451) & 16\%(88) \\
        None & 13\% (173) & - & - \\
        GPT-3 & 45\% (559) & 33\%(186) & 66\% (373)\\\hline\hline
        DPR+FiD & 46\% (578) & 95\% (547) & 5\% (31) \\
        None & 9\% (109) & - & -  \\
        Human  & \textbf{74\% (931)$\dagger$} & 94\% (879) & 6\% (52) \\\hline
    \end{tabular}
    \caption{Pair-wise Preference task results on a sample of 250 examples with 5 annotations. Baseline (DPR+FiD) is compared with GPT-3 variants, and human responses. Users can select both models or none. Total number of annotations per model is in parentheses. Last two columns denote a breakdown of selected responses that were faithful, or unfaithful. $\dagger$ indicates stat. sig. against the baseline using $\chi^2$ goodness of fit ($p < .05$).}
    \label{tab:preferences}
\end{table}
%


\section{Study of Trust}


So far we have established that lexically aligned responses coming from \mbox{GPT-3} variants are not necessarily faithful. The surface form seems to negatively affect users' preferences, obviating their need to check the supporting source, and creating a risk of placing trust to an imperfect system. With this experiment, we investigate a more general trend between linguistic phenomena and user trust.

\shortpar{Human Evaluation Experiment} Annotators are presented with the dialogue only, and are asked to choose the response they trusted more from two possible responses, or none. 
Going beyond just lexical alignment, we selected 15 pairs of responses\footnote{Note that we select only faithful responses, explicitly informing participants.}, for every linguistic phenomenon in Section~\ref{sec:dialogue-phenomena}. We modified responses to ensure each specific phenomenon was the only difference between them. We  collected 20 preferences for each response pair. 

\shortpar{Results} Table~\ref{tab:trust-human} shows that annotators trusted responses with high lexical alignment significantly more than those with low lexical alignment. 
Interestingly, they trusted significantly more short answers than fragments, and preferred responses that did not present pronouns. This is in contrast to literature \cite{https://doi.org/10.1111/cogs.12225}, which primarily focused on human-to-human interactions; this could be down to people talking to a system (vs. a human), seeking stronger forms of evidence such as lexical alignment. 
Notably, the combination of the preferred presence and absence of phenomena aligns well with their calculated occurrences in the GPT-3 variants' responses (Table~\ref{tab:dialogue-phenomena}).


%
\begin{table}[]
    \small
    \centering
    \begin{tabular}{r|c}\hline
        \textbf{Linguistic phenomena} & \textbf{Trust}\\\hline         
        High Lexical Alignment & \textbf{58\%$\dagger$}  \\
        None & 10\%  \\
        Low Lexical Alignment & 32\%  \\\hline\hline
        Pronouns & 31\%  \\
        None & 19\%  \\
        No Pronouns & \textbf{49\%$\dagger$}  \\\hline\hline
        Short answer & \textbf{66\%$\dagger$} \\
        None & 7\%  \\
        Fragment & 26\% \\\hline
    \end{tabular}
    \caption{Human Evaluation experiment on Trust for various linguistic phenomena. High/Low lexical alignment threshold is set to 0.5, based on recall. $\dagger$ denotes pair-wise stat. sig. using $\chi^2$ goodness of fit ($p<.05$).}
    \label{tab:trust-human}
\end{table}

\section{Conclusions}
We investigated the performance of different models on the task of OCQA, measuring faithfulness and lexical phenomena. Automatic metrics highlighted how GPT-3 variants are less faithful than DPR+FiD, as confirmed by annotators in the faithfulness judgment task. We conducted a study on conversational grounding phenomena and a preference task, whose significant results demonstrated an effect of surface form in human preferences towards the more conversational GPT-3, even when unfaithful. Another experiment confirmed trust as being effected by high lexical alignment. 

\section*{Limitations}

This work is constrained by the number of grounding phenomena analyzed, which is limited by the dataset domain and their straightforward automatic computation. We only focused on lexical alignment, the use of ellipsis (fragments) and pronouns, disregarding other phenomena such as repairs (e.g. asking for confirmation or clarification)~\cite{purver2003means}, among others. 

With respect to the linguistic phenomena, we simplified the calculation of the lexical alignment by regarding only the last two turns of a conversation (the user question and the system response). In this manner, we omitted the dynamic convergence over several turns~\cite{mills-healey-2008-semantic}. It should be noted though that this was decided based on manual observation of examples, the majority of which exhibited lexical alignment in the last two turns only. This could be a limitation of the OCQA domain, and/or a bias of the TopiOCQA dataset.

Another limitation is that the form of crowd-sourcing experiments we performed are mostly diagnostic of certain conditions on a given dataset, and does not reflect more organic real-use cases. An ideal setup would be to collect whole dialogues in the form of an extrinsic evaluation, which would be more costly to perform.




\section*{Ethics Statement}
\paragraph{Dual Use} Our results highlight a possible misuse scenario, where verbally fluent but factually incorrect text generated by models, such as GPT-3, is more convincing to users than text by models which are more faithful to the input rationale. This blind trust could be exploited to convince users of e.g.\ fake news, for example by generating more lexically aligned text. 

\paragraph{Human data} The methodology of this paper heavily relies on human data collection using crowd-sourcing. Workers were allowed to complete a maximum of 40 HiTs across annotations. They were payed 0.29\$ per HiT for the preference task, while 0.20\$ per HiT for the study on trust. Annotators come from Australia, Canada, New Zeland, United Kingdom and United States. A total of 38 annotators were involved in the study of trust, and 115 were involved in the Preference task. Data collected using AMT are fully anonymized per the providers specifications.

\paragraph{Use of TopiOCQA} We obtained the dataset through the public domain and do not intend to release part, or whole of it separately without the prior consent of its authors. We assume the authors have taken precautions against offensive content.

\section*{Acknowledgements}
We would like to particularly thank Oliver Lemon for the discussions on the linguistic phenomena in conversation and trust. We also appreciate the valuable feedback we received by the rest of the technical team at Alana AI at various stages of the paper. Finally, we would like to thank the anonymous reviewers and annotators for the human evaluation.

\bibliography{anthology,custom}

\begin{thebibliography}{27}
\expandafter\ifx\csname natexlab\endcsname\relax\def\natexlab#1{#1}\fi

\bibitem[{Adlakha et~al.(2022)Adlakha, Dhuliawala, Suleman, de~Vries, and Reddy}]{adlakha-etal-2022-topiocqa}
Vaibhav Adlakha, Shehzaad Dhuliawala, Kaheer Suleman, Harm de~Vries, and Siva Reddy. 2022.
\newblock \href {https://doi.org/10.1162/tacl_a_00471} {{T}opi{OCQA}: Open-domain conversational question answering with topic switching}.
\newblock \emph{Transactions of the Association for Computational Linguistics}, 10:468--483.

\bibitem[{Bai et~al.(2022)Bai, Jones, Ndousse, Askell, Chen, DasSarma, Drain, Fort, Ganguli, Henighan et~al.}]{bai2022training}
Yuntao Bai, Andy Jones, Kamal Ndousse, Amanda Askell, Anna Chen, Nova DasSarma, Dawn Drain, Stanislav Fort, Deep Ganguli, Tom Henighan, et~al. 2022.
\newblock Training a helpful and harmless assistant with reinforcement learning from human feedback.
\newblock \emph{arXiv preprint arXiv:2204.05862}.

\bibitem[{Bickmore et~al.(2021)Bickmore, {\'O}lafsson, and O'Leary}]{Bickmore2021:medicalTrust}
Timothy~W Bickmore, Stef{\'a}n {\'O}lafsson, and Teresa~K O'Leary. 2021.
\newblock \href {https://doi.org/10.2196/30704} {Mitigating patient and consumer safety risks when using conversational assistants for medical information: Exploratory mixed methods experiment}.
\newblock \emph{J Med Internet Res}, 23(11):e30704.

\bibitem[{Bossens and Evers(2022)}]{Bossens2022:HRIgrounding}
David~M. Bossens and Christine Evers. 2022.
\newblock \href {https://doi.org/10.48550/ARXIV.2209.02066} {Trust in language grounding: a new ai challenge for human-robot teams}.

\bibitem[{Branigan et~al.(2010)Branigan, Pickering, Pearson, and McLean}]{branigan2010linguistic}
Holly~P Branigan, Martin~J Pickering, Jamie Pearson, and Janet~F McLean. 2010.
\newblock Linguistic alignment between people and computers.
\newblock \emph{Journal of pragmatics}, 42(9):2355--2368.

\bibitem[{Brown et~al.(2020)Brown, Mann, Ryder, Subbiah, Kaplan, Dhariwal, Neelakantan, Shyam, Sastry, Askell, Agarwal, Herbert-Voss, Krueger, Henighan, Child, Ramesh, Ziegler, Wu, Winter, Hesse, Chen, Sigler, Litwin, Gray, Chess, Clark, Berner, McCandlish, Radford, Sutskever, and Amodei}]{NEURIPS2020_1457c0d6}
Tom Brown, Benjamin Mann, Nick Ryder, Melanie Subbiah, Jared~D Kaplan, Prafulla Dhariwal, Arvind Neelakantan, Pranav Shyam, Girish Sastry, Amanda Askell, Sandhini Agarwal, Ariel Herbert-Voss, Gretchen Krueger, Tom Henighan, Rewon Child, Aditya Ramesh, Daniel Ziegler, Jeffrey Wu, Clemens Winter, Chris Hesse, Mark Chen, Eric Sigler, Mateusz Litwin, Scott Gray, Benjamin Chess, Jack Clark, Christopher Berner, Sam McCandlish, Alec Radford, Ilya Sutskever, and Dario Amodei. 2020.
\newblock \href {https://proceedings.neurips.cc/paper/2020/file/1457c0d6bfcb4967418bfb8ac142f64a-Paper.pdf} {Language models are few-shot learners}.
\newblock In \emph{Advances in Neural Information Processing Systems}, volume~33, pages 1877--1901. Curran Associates, Inc.

\bibitem[{Clark and Brennan(1991)}]{clark1991grounding}
Herbert~H Clark and Susan~E Brennan. 1991.
\newblock Grounding in communication.

\bibitem[{Dziri et~al.(2022{\natexlab{a}})Dziri, Kamalloo, Milton, Zaiane, Yu, Ponti, and Reddy}]{https://doi.org/10.48550/arxiv.2204.10757}
Nouha Dziri, Ehsan Kamalloo, Sivan Milton, Osmar Zaiane, Mo~Yu, Edoardo~M. Ponti, and Siva Reddy. 2022{\natexlab{a}}.
\newblock \href {https://doi.org/10.48550/ARXIV.2204.10757} {Faithdial: A faithful benchmark for information-seeking dialogue}.

\bibitem[{Dziri et~al.(2022{\natexlab{b}})Dziri, Milton, Yu, Zaiane, and Reddy}]{dziri-etal-2022-origin}
Nouha Dziri, Sivan Milton, Mo~Yu, Osmar Zaiane, and Siva Reddy. 2022{\natexlab{b}}.
\newblock \href {https://doi.org/10.18653/v1/2022.naacl-main.387} {On the origin of hallucinations in conversational models: Is it the datasets or the models?}
\newblock In \emph{Proceedings of the 2022 Conference of the North American Chapter of the Association for Computational Linguistics: Human Language Technologies}, pages 5271--5285, Seattle, United States. Association for Computational Linguistics.

\bibitem[{Eshghi and Healey(2016)}]{https://doi.org/10.1111/cogs.12225}
Arash Eshghi and Patrick G.~T. Healey. 2016.
\newblock \href {https://doi.org/https://doi.org/10.1111/cogs.12225} {Collective contexts in conversation: Grounding by proxy}.
\newblock \emph{Cognitive Science}, 40(2):299--324.

\bibitem[{Fernandez and Ginzburg(2002)}]{fernandez-ginzburg-2002-non-sentential}
Raquel Fernandez and Jonathan Ginzburg. 2002.
\newblock \href {https://doi.org/10.3115/1118121.1118124} {Non-sentential utterances in dialogue: A: Corpus-based study}.
\newblock In \emph{Proceedings of the Third {SIG}dial Workshop on Discourse and Dialogue}, pages 15--26, Philadelphia, Pennsylvania, USA. Association for Computational Linguistics.

\bibitem[{Glaese et~al.(2022)Glaese, McAleese, Tr{\k{e}}bacz, Aslanides, Firoiu, Ewalds, Rauh, Weidinger, Chadwick, Thacker et~al.}]{glaese2022improving}
Amelia Glaese, Nat McAleese, Maja Tr{\k{e}}bacz, John Aslanides, Vlad Firoiu, Timo Ewalds, Maribeth Rauh, Laura Weidinger, Martin Chadwick, Phoebe Thacker, et~al. 2022.
\newblock Improving alignment of dialogue agents via targeted human judgements.
\newblock \emph{arXiv preprint arXiv:2209.14375}.

\bibitem[{Honovich et~al.(2021)Honovich, Choshen, Aharoni, Neeman, Szpektor, and Abend}]{honovich-etal-2021-q2}
Or~Honovich, Leshem Choshen, Roee Aharoni, Ella Neeman, Idan Szpektor, and Omri Abend. 2021.
\newblock \href {https://doi.org/10.18653/v1/2021.emnlp-main.619} {$q^{2}$: {E}valuating factual consistency in knowledge-grounded dialogues via question generation and question answering}.
\newblock In \emph{Proceedings of the 2021 Conference on Empirical Methods in Natural Language Processing}, pages 7856--7870, Online and Punta Cana, Dominican Republic. Association for Computational Linguistics.

\bibitem[{Izacard and Grave(2021)}]{izacard-grave-2021-leveraging}
Gautier Izacard and Edouard Grave. 2021.
\newblock \href {https://doi.org/10.18653/v1/2021.eacl-main.74} {Leveraging passage retrieval with generative models for open domain question answering}.
\newblock In \emph{Proceedings of the 16th Conference of the European Chapter of the Association for Computational Linguistics: Main Volume}, pages 874--880, Online. Association for Computational Linguistics.

\bibitem[{Karpukhin et~al.(2020)Karpukhin, Oguz, Min, Lewis, Wu, Edunov, Chen, and Yih}]{karpukhin-etal-2020-dense}
Vladimir Karpukhin, Barlas Oguz, Sewon Min, Patrick Lewis, Ledell Wu, Sergey Edunov, Danqi Chen, and Wen-tau Yih. 2020.
\newblock \href {https://doi.org/10.18653/v1/2020.emnlp-main.550} {Dense passage retrieval for open-domain question answering}.
\newblock In \emph{Proceedings of the 2020 Conference on Empirical Methods in Natural Language Processing (EMNLP)}, pages 6769--6781, Online. Association for Computational Linguistics.

\bibitem[{Lin(2004)}]{Lin2004}
Chin-Yew Lin. 2004.
\newblock \href {https://aclanthology.org/W04-1013} {{ROUGE}: A package for automatic evaluation of summaries}.
\newblock In \emph{Text Summarization Branches Out}, pages 74--81, Barcelona, Spain. Association for Computational Linguistics.

\bibitem[{Linnemann and Jucks(2018)}]{Linnemann2018:lexicalAlignmentTrust}
Gesa~Alena Linnemann and Regina Jucks. 2018.
\newblock \href {https://doi.org/10.1093/iwc/iwy005} {{‘Can I Trust the Spoken Dialogue System Because It Uses the Same Words as I Do?’—Influence of Lexically Aligned Spoken Dialogue Systems on Trustworthiness and User Satisfaction}}.
\newblock \emph{Interacting with Computers}, 30(3):173--186.

\bibitem[{Mills and Healey(2008)}]{mills-healey-2008-semantic}
Gregory Mills and Pat Healey. 2008.
\newblock \href {https://aclanthology.org/W08-0106} {Semantic negotiation in dialogue: the mechanisms of alignment}.
\newblock In \emph{Proceedings of the 9th {SIG}dial Workshop on Discourse and Dialogue}, pages 46--53, Columbus, Ohio. Association for Computational Linguistics.

\bibitem[{Papineni et~al.(2002)Papineni, Roukos, Ward, and Zhu}]{PapineniEtAl2002}
Kishore Papineni, Salim Roukos, Todd Ward, and Wei-Jing Zhu. 2002.
\newblock \href {https://doi.org/10.3115/1073083.1073135} {{B}leu: a method for automatic evaluation of machine translation}.
\newblock In \emph{Proceedings of the 40th Annual Meeting of the Association for Computational Linguistics}, pages 311--318, Philadelphia, Pennsylvania, USA. Association for Computational Linguistics.

\bibitem[{Pickering and Garrod(2004)}]{pickering2004toward}
Martin~J Pickering and Simon Garrod. 2004.
\newblock Toward a mechanistic psychology of dialogue.
\newblock \emph{Behavioral and brain sciences}, 27(2):169--190.

\bibitem[{Purver et~al.(2003)Purver, Ginzburg, and Healey}]{purver2003means}
Matthew Purver, Jonathan Ginzburg, and Patrick Healey. 2003.
\newblock On the means for clarification in dialogue.
\newblock In \emph{Current and new directions in discourse and dialogue}, pages 235--255. Springer.

\bibitem[{Qi et~al.(2020)Qi, Zhang, Zhang, Bolton, and Manning}]{QiEtAl2020}
Peng Qi, Yuhao Zhang, Yuhui Zhang, Jason Bolton, and Christopher~D. Manning. 2020.
\newblock \href {https://doi.org/10.18653/v1/2020.acl-demos.14} {{S}tanza: A python natural language processing toolkit for many human languages}.
\newblock In \emph{Proceedings of the 58th Annual Meeting of the Association for Computational Linguistics: System Demonstrations}, pages 101--108, Online. Association for Computational Linguistics.

\bibitem[{Rashkin et~al.(2021{\natexlab{a}})Rashkin, Nikolaev, Lamm, Aroyo, Collins, Das, Petrov, Tomar, Turc, and Reitter}]{rashkin2021measuring}
Hannah Rashkin, Vitaly Nikolaev, Matthew Lamm, Lora Aroyo, Michael Collins, Dipanjan Das, Slav Petrov, Gaurav~Singh Tomar, Iulia Turc, and David Reitter. 2021{\natexlab{a}}.
\newblock Measuring attribution in natural language generation models.
\newblock \emph{arXiv preprint arXiv:2112.12870}.

\bibitem[{Rashkin et~al.(2021{\natexlab{b}})Rashkin, Reitter, Tomar, and Das}]{rashkin-etal-2021-increasing}
Hannah Rashkin, David Reitter, Gaurav~Singh Tomar, and Dipanjan Das. 2021{\natexlab{b}}.
\newblock \href {https://doi.org/10.18653/v1/2021.acl-long.58} {Increasing faithfulness in knowledge-grounded dialogue with controllable features}.
\newblock In \emph{Proceedings of the 59th Annual Meeting of the Association for Computational Linguistics and the 11th International Joint Conference on Natural Language Processing (Volume 1: Long Papers)}, pages 704--718, Online. Association for Computational Linguistics.

\bibitem[{Shuster et~al.(2021)Shuster, Poff, Chen, Kiela, and Weston}]{shuster2021retrieval}
Kurt Shuster, Spencer Poff, Moya Chen, Douwe Kiela, and Jason Weston. 2021.
\newblock Retrieval augmentation reduces hallucination in conversation.
\newblock \emph{arXiv preprint arXiv:2104.07567}.

\bibitem[{Thoppilan et~al.(2022)Thoppilan, De~Freitas, Hall, Shazeer, Kulshreshtha, Cheng, Jin, Bos, Baker, Du et~al.}]{thoppilan2022lamda}
Romal Thoppilan, Daniel De~Freitas, Jamie Hall, Noam Shazeer, Apoorv Kulshreshtha, Heng-Tze Cheng, Alicia Jin, Taylor Bos, Leslie Baker, Yu~Du, et~al. 2022.
\newblock Lamda: Language models for dialog applications.
\newblock \emph{arXiv preprint arXiv:2201.08239}.

\bibitem[{Zhang et~al.(2019)Zhang, Kishore, Wu, Weinberger, and Artzi}]{zhang2019bertscore}
Tianyi Zhang, Varsha Kishore, Felix Wu, Kilian~Q Weinberger, and Yoav Artzi. 2019.
\newblock Bertscore: Evaluating text generation with bert.
\newblock \emph{arXiv preprint arXiv:1904.09675}.

\end{thebibliography}
\bibliographystyle{acl_natbib}

\newpage

\appendix



\section{Additional Human Evaluation Results}
\label{app:additional_human_results}

\shortpar{Majority Agreement Results} 

\noindent Following \citet{glaese2022improving} we computed the majority agreement for each item, i.e., 5 and 20 annotations per item for the preference and trust studies, respectively. Tables~\ref{tab:prefs_majority_agreement} and \ref{tab:trust_majority_agreement} summarize the results. Similar to \citet{glaese2022improving} there are cases when agreement is quite low, which is an interesting avenue for future work.

\shortpar{Qualitative Analysis of Feedback} 

\noindent Next, we conducted a simple qualitative analysis regarding how often annotators looked at the grounded document during the Preference Task. 286 out of 2170 feedback responses explicitly refer to the document to justify the preference expressed. Interestingly, There are in total 558 responses where GPT-3 variants were preferred over the baseline, of which only 27 (4\%) refer to the document. In contrast, there are 359 of which 76 refer to the document (21\%) when the baseline is preferred. Overall, feedback suggests that GPT-3 responses were mostly preferred due to other factors, such as the amount and variety of information, and conversational style.

\begin{table}[t]
    \centering
    \small
    \begin{tabular}{r|c}
        \textbf{Baseline vs } & \textbf{Agreement} \\\hline
        DPR+GPT-3 & 86.4\% \\
        GPT-3 & 77.6\% \\
        Human & 90\%
    \end{tabular}
    \caption{Majority Agreement per item (5 annotations) for the Preference Task between the Baseline (DPR+FiD) and models. Each row denotes majority reached at the corresponding \% of the times.}
    \label{tab:prefs_majority_agreement}
\end{table}

\begin{table}[t]
    \centering
    \small
    \begin{tabular}{r|c}
        \textbf{Phenomenon} & \textbf{Agreement} \\\hline
        Lexical Alignment & 80\% \\
        Pronouns & 53\% \\
        Fragment & 86\%
    \end{tabular}
    \caption{Majority Agreement per item (20 annotations) for the Study of Trust across the different linguistic phenomena examined in this work. Each row denotes majority reached at the corresponding \% of the times.}
    \label{tab:trust_majority_agreement}
\end{table}

\section{Human Evaluation Instructions and Interfaces}
\label{sec:human-evaluation-appendix}

\subsection{Faithfulness Judgment Task}\label{sec:human-evaluation-appendix-faithfulness}
\begin{figure*}
    \centering
    \includegraphics[scale=0.5]{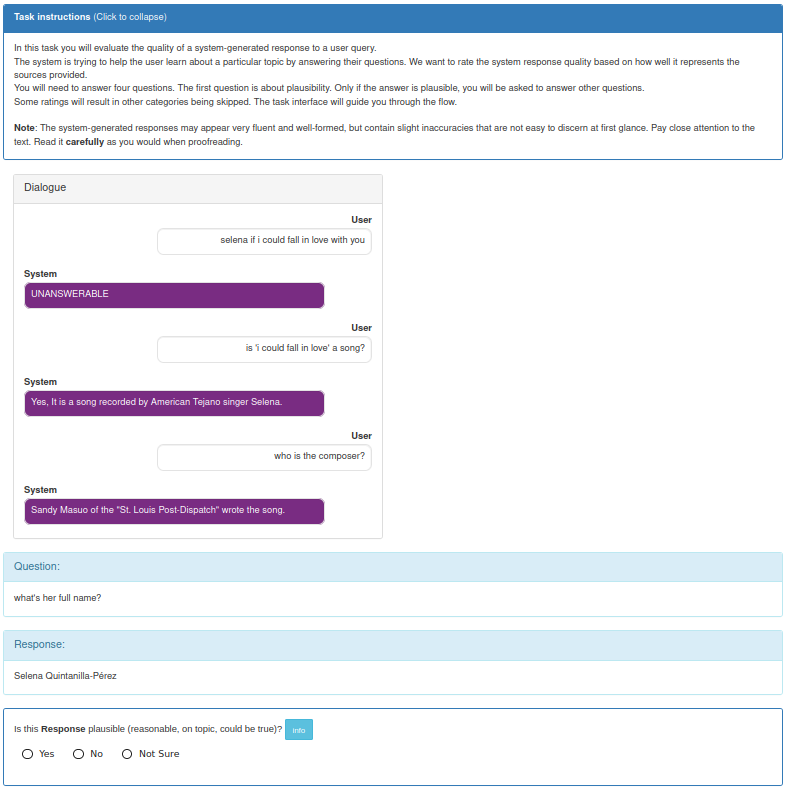}
    \caption{Interface used to collect faithfulness. The annotator is asked to answer the question about plausibility of the response first, without looking at the document. The annotation stops at this point if the response is not plausible.}
    \label{fig::faith-layout-1}
\end{figure*}

\begin{figure*}
    \centering
    \includegraphics[scale=0.5]{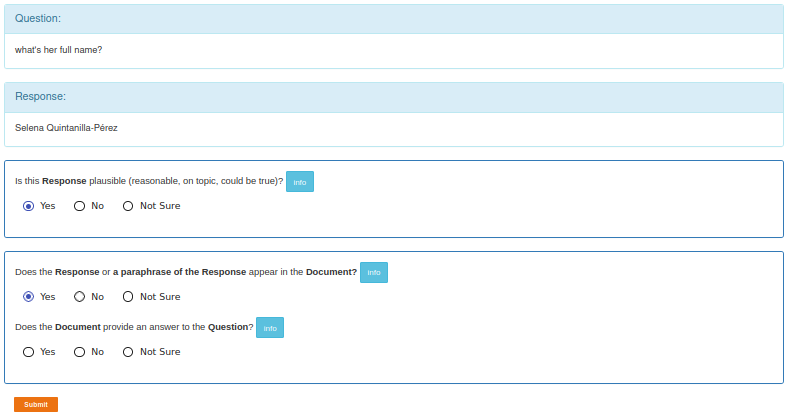}
    \caption{Interface used to collect faithfulness. The annotator has now access to the document and can annotate.}
    \label{fig::faith-layout-2}
\end{figure*}

Figures~\ref{fig::faith-layout-1} and \ref{fig::faith-layout-2} illustrate the user interface implemented for the plausibility and faithfulness sub-tasks, respectively.

\shortpar{Task Instructions:} 

\noindent In this task you will evaluate the quality of a system-generated response to a user query. The system is trying to help the user learn about a particular topic by answering their questions. We want to rate the system response quality based on how well it represents the sources provided. You will need to answer four questions. The first question is about plausibility. Only if the answer is plausible, you will be asked to answer other questions. Some ratings will result in other categories being skipped. The task interface will guide you through the flow.

\noindent \textbf{Note}: The system-generated responses may appear very fluent and well-formed, but contain slight inaccuracies that are not easy to discern at first glance. Pay close attention to the text. Read it carefully as you would when proofreading.

\subsection{Preference Task}\label{sec:human-evaluation-appendix-preference}
\begin{figure*}
    \centering
    \includegraphics[scale=0.5]{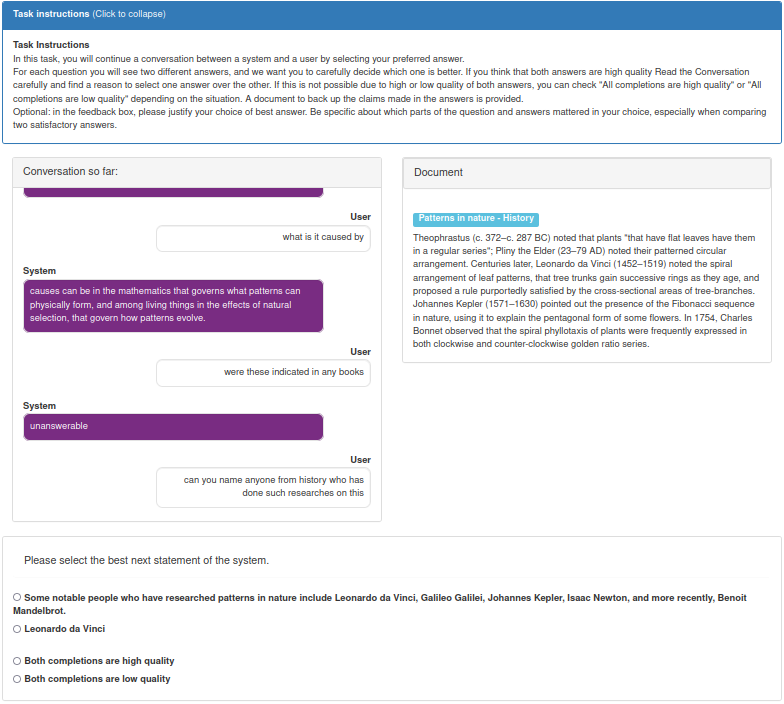}
    \caption{Interface used to collect the human evaluation for preferences}
    \label{fig::prefs-layout}
\end{figure*}

Figure~\ref{fig::prefs-layout} depicts the interface for the preference task in the context of the dialogue and gold passage.

\shortpar{Task Instructions:} 

\noindent In this task, you will continue a conversation between a system and a user by selecting your preferred answer. For each question you will see two different answers, and we want you to carefully decide which one is better. Read the Conversation carefully and find a reason to select one answer over the other. If this is not possible due to high or low quality of both answers, you can check "All completions are high quality" or "All completions are low quality" depending on the situation. A document to back up the claims made in the answers is provided.

\noindent \textbf{Optional}: in the feedback box, please justify your choice of best answer. Be specific about which parts of the question and answers mattered in your choice, especially when comparing two satisfactory answers.

\subsection{Study of Trust}\label{sec:human-evaluation-appendix-trust}
\begin{figure*}
    \centering
    \includegraphics[scale=0.5]{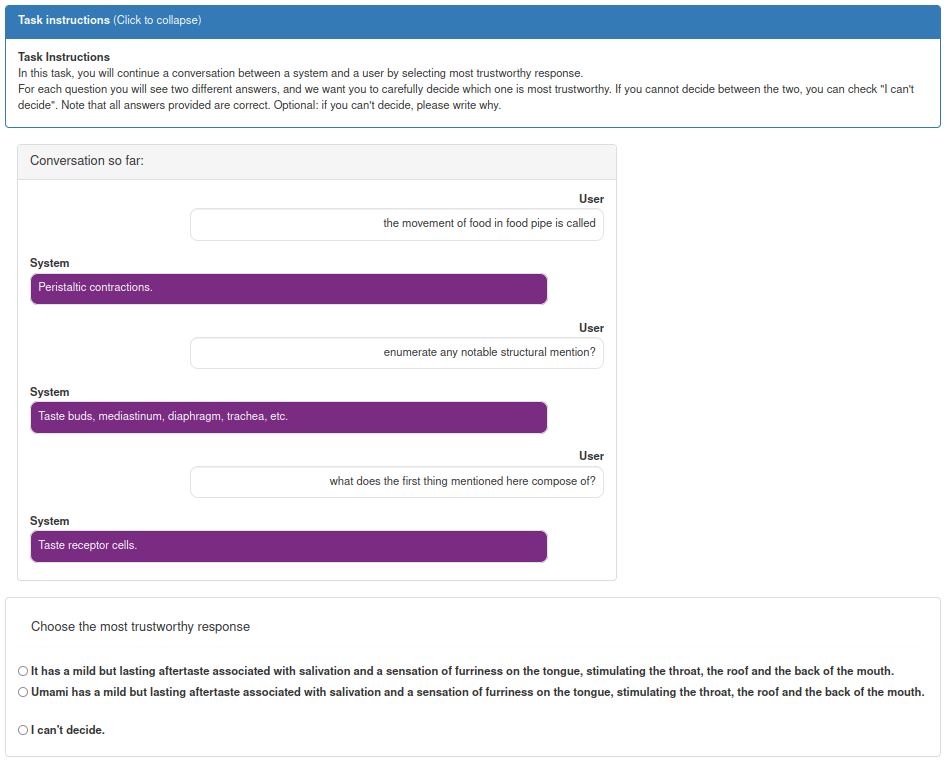}
    \caption{Interface used to collect the human evaluation for the study of trust}
    \label{fig::trust-layout}
\end{figure*}

Figure~\ref{fig::trust-layout} shows a screenshot of the trust task given the dialogue only.

\shortpar{Task Instructions:} 

\noindent In this task, you will continue a conversation between a system and a user by selecting most trustworthy response. For each question you will see two different answers, and we want you to carefully decide which one is most trustworthy. If you cannot decide between the two, you can check "I can't decide". Note that all answers provided are correct. 

\noindent \textbf{Optional}: if you can't decide, please write why. 

%

\end{document}